\documentclass[runningheads,a4paper]{llncs}

\usepackage{amssymb}
\usepackage{amsmath}
\usepackage{graphicx}
\usepackage[table,pdftex]{xcolor}
\usepackage{xspace}
\usepackage[colorlinks=true,citecolor=blue,urlcolor=black]{hyperref}
\usepackage{lineno}
\usepackage{subfig}
\usepackage[colorinlistoftodos]{todonotes}
\usepackage{wrapfig}
\usepackage[misc]{ifsym}
\usepackage{amsfonts}
\usepackage{booktabs}
\newcommand{\vect}[1]{\boldsymbol{#1}} 

\usepackage{algorithm}
\usepackage{algpseudocode}

\begin{document}

\mainmatter  
\title{OCT segmentation: Integrating open parametric contour model of the retinal layers and shape constraint to the Mumford-Shah functional}
\titlerunning{Shape-based variational Mumford-Shah model}

\newcommand{\corrauth}{\textsuperscript{(\Letter)}}
\author{Jinming Duan$^{1}$\corrauth, Weicheng Xie$^{2}$, Ryan Wen Liu$^{3}$, Christopher Tench$^{4}$,\\ Irene Gottlob$^{5}$, Frank Proudlock$^{5}$, Li Bai$^{1}$}
\authorrunning{J Duan, et al.}
\institute{$^1$School of Computer Science, University of Nottingham, UK\\
$^2$School of Computer Science and Software Engineering, Shenzhen University, China\\
$^3$School of Navigation, Wuhan University of Technology, China\\
$^4$School of Medicine, University of Nottingham, UK \\
$^5$Ophthalmology Department, University of Leicester, UK\\
\email{Jiming.Duan@nottingham.ac.uk}\\}

\maketitle



\begin{abstract}

In this paper, we propose a novel retinal layer boundary model for segmentation of optical coherence tomography (OCT) images. The retinal layer boundary model consists of 9 open parametric contours representing the 9 retinal layers in OCT images. An intensity-based Mumford-Shah (MS) variational functional is first defined to evolve the retinal layer boundary model to segment the 9 layers simultaneously. By making use of the normals of open parametric contours, we construct equal sized adjacent narrowbands that are divided by each contour. Regional information in each narrowband can thus be integrated into the MS energy functional such that its optimisation is robust against different initialisations. A statistical prior is also imposed on the shape of the segmented parametric contours for the functional. As such, by minimising the MS energy functional the parametric contours can be driven towards the true boundaries of retinal layers, while the similarity of the contours with respect to training OCT shapes is preserved. Experimental results on real OCT images demonstrate that the method is accurate and robust to low quality OCT images with low contrast and high-level speckle noise, and it outperforms the recent geodesic distance based method for segmenting 9 layers of the retina in OCT images.

\end{abstract}


\section{Introduction}

Optical coherence tomography (OCT) image segmentation to detect retinal layer boundaries is a fundamental procedure for diagnosing and monitoring the progression of retinal and optical nerve diseases. There exist rich literature on approaches for automatic and semi-automatic OCT image segmentation. Common methods include deformable models \cite{rossant2015parallel,chan2001active}, graph-based and geodesic distance methods {\cite{chiu2010automatic,duan2017automated}, statistical shape and appearance models \cite{Xie2018Open,cootes2000introduction}, etc. Very recently, deep neural networks are becoming increasingly popular for OCT segmentation, demonstrating excellent performance  \cite{roy2017relaynet,fang2017automatic}. However, these deep learning methods usually require the networks to be sufficiently deep to learn all appearance and shape variations of the retinal layers from annotated training sets. Therefore the training set has to be very large and rich to prevent the over-fitting. Large annotated training sets are however difficult to obtain. Existing OCT segmentation algorithms also tend to segment individual retinal layers separately. This form of analysis often fails when there is uncertainty in the image, especially some retinal layers are often difficult to see or missing in OCT images.


We believe that OCT segmentation is more effective by incorporating anatomical shape of the retinal layers and their spatial relations. As such, in this paper we propose a shape-based variational Mumford-Shah (MS) functional for segmentation of up to 9 retinal layer boundaries in OCT images, using only a small training set. We make three distinct contributions to OCT segmentation:

\begin{itemize}
\item	We introduce a new piecewise constant variational MS functional to evolve a pre-defined retinal layer boundary model for OCT image segmentation. It has a region-based data fidelity term and a hybridised first and second regularisation term. The pre-defined retinal layer boundary model consists of 9 retinal layer boundaries, each is an open explicit parametric contour represented by a set of control points. We then construct two narrowbands around each open contour, within which region-based information is derived to aid contour evolution. We show that by incorporating a retinal layer boundary model our method can segment 9 retinal layers simultaneously, and by utilising regional information, the proposed method has a large convergence range and is robust to initialisation.

\item	We introduce to the MS functional a shape constraint learnt from a set of training OCT shapes. We then apply the principal component analysis (PCA) to derive the statistical distribution from the training shapes as well as the resulting irregular contours evolved directly from the MS functional. In this way, the irregular contours are restricted to a manifold of familiar shapes and thereby can be pulled back to appropriate positions to allow a faster convergence.

\item	We apply the proposed method to real OCT dataset acquired from healthy subjects, and demonstrate that the proposed method outperforms the state-of-the-art methods.
\end{itemize}


\section{Methodology}
%
\subsection{Intensity-based variational MS functional}
We start with a new intensity-based MS segmentation functional, and then apply a learnt shape constraint to the functional. We define a retinal layer boundary model as having 9 retinal layers, each of which is an explicit parametric contour $C$. Based on \cite{cremers2002diffusion}, we propose a new piecewise constant MS functional for each of the 9 retinal layers
\begin{equation} \label{eq:ourModel}
E\left( {\left\{ {{u_i}} \right\},C} \right)  = \sum\limits_i {\int_{{\Omega_i}} {{{\left( {f - {u_i}} \right)}^2}} dx + \frac{\alpha }{2}\int_0^1 {{{\left| {{C_s}\left( s \right)} \right|}^2}ds}  + \frac{\beta }{2}\int_0^1 {{{\left| {{C_{ss}}\left( s \right)} \right|}^2}ds}},
\end{equation}
where $f$ is the input image and $u_i$ are the mean grey value of region $\Omega_i$. The first regional energy term follows the Chan-Vese model \cite{chan2001active,duan2014some}. In the second term, $C_s$ is the first order derivative of the curve $C$ with respect to the arc length $s$ normalized into the region $[0,1]$, and $C_{ss}$ is the second order derivative. $\alpha$ and $\beta$ are two regularisation coefficients. To segment a retinal layer in OCT, $i$ has two values, namely, $1,2$. The contour regularisation (last two terms) combines the first and second order derivatives, preventing the contour from bending by introducing elasticity and stiffness to the contour. Each control point of the contour thus can be more equidistant or centred between its neighbourhood points. This makes the functional stable for numerical calculation. $C$ is defined as a parametric contour of a set of control points
\begin{equation} \label{eq:ctrpitvector}
C(s) = {(\vect{x},\vect{y})^T}.
\end{equation}
Here $\vect{x} = ({x_1}, \cdots, {x_M})$, $\vect{y} = ({y_1}, \cdots, {y_M})$ are the coordinates of the $M$ control points to represent one retinal layer boundary. The start and end points are $C(0)=(x_1,y_1)^T$ and $C(1)=(x_M,y_M)^T$, respectively.

\vspace{15pt}
\begin{figure}
\vspace{-20pt}
\centering
\includegraphics[width=0.7\textwidth]{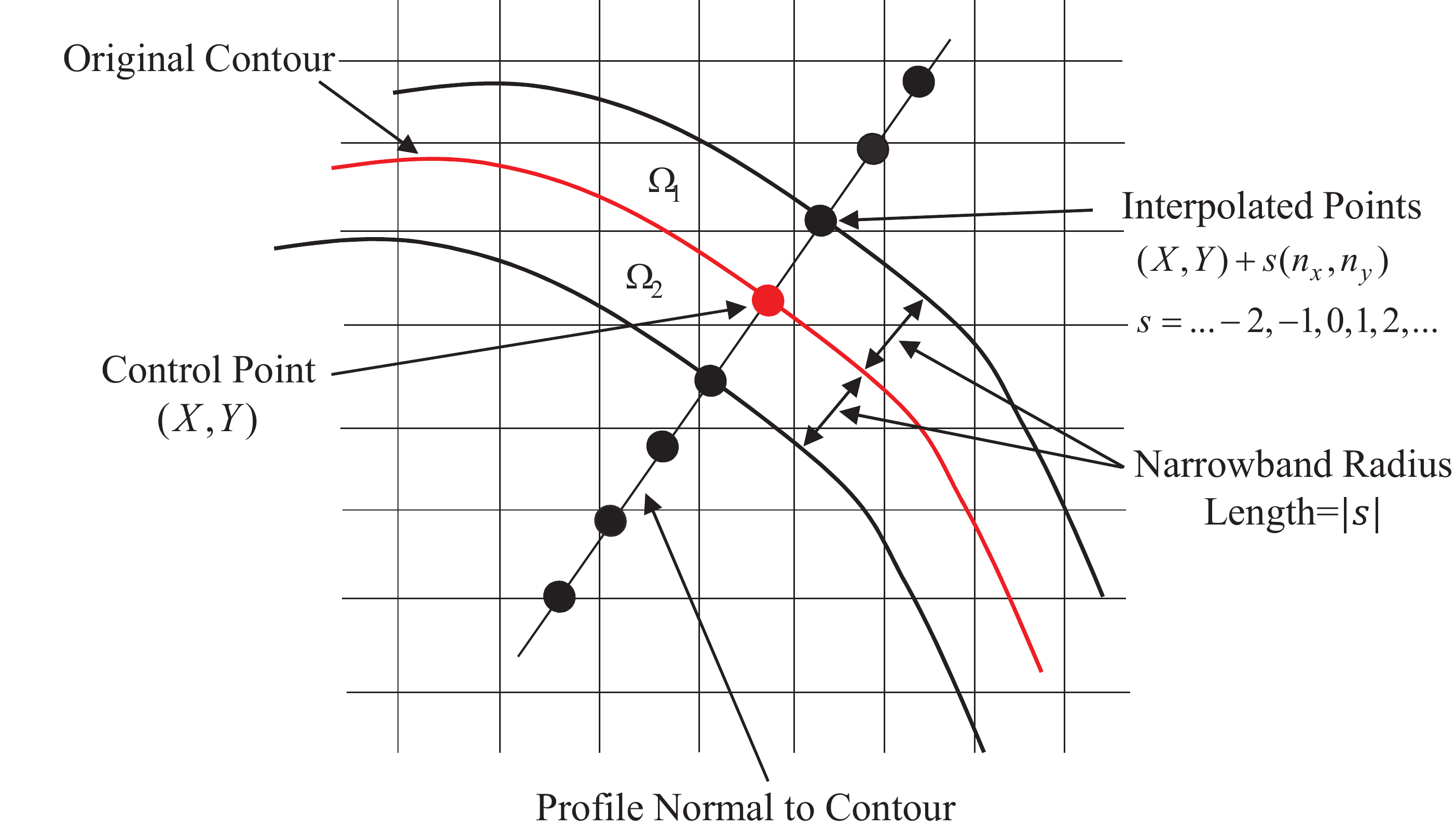}
\caption{A strategy to construct a narrowband around an open contour. In the narrowband region information can be utilised.}
\label{fig:narrowBand}
\end{figure}

Variational methods taking into account regional information are generally more robust against noisy and initialisation. In OCT segmentation, we however require the parametric contour $C$ to be an open curve \cite{Xie2018Open}. In this case, $C(0)$ is no longer equal to $C(1)$. This brings difficulties in estimating parameter $(u_1, u_2)$ in (\ref{eq:ourModel}) due to the vanished image regions $\Omega_1$ and $\Omega_2$. To circumvent this limitation, we propose to construct a narrowband around the open contour $C$ using the method illustrated in Fig~\ref{fig:narrowBand}: 

\begin{itemize}

\item For each control point on $C$, such as the red dot in Fig~\ref{fig:narrowBand}, we compute its normal using the neighbourhood control points; 

\item We define a narrowband radius {$|s|$}, and interpolate new points, as represented by the black dots in Fig~\ref{fig:narrowBand}, along the two normal directions of the original control points; 

\item We determine those pixels that have fallen in the narrowband between the two contours in Fig~\ref{fig:narrowBand} and afterwards compute $(u_1, u_2)$ as in (\ref{eq:statisticsCal}). 

\end{itemize}

\subsection{Minimisation of MS by gradient descent}
We minimise the proposed functional (\ref{eq:ourModel}) with respect to both the parameter $(u_1,u_2)$ and the contour $C$, where $(u_1,u_2)$ denote the mean grey values on both sides of contour curve $C$ and have a simple closed-form
\begin{equation} \label{eq:statisticsCal}
{u_i} = \frac{1}{{\left| {{\Omega _i}} \right|}}\int_{{\Omega _i}} {fdx}, \;\;\; i=1, 2,
\end{equation}
where $\Omega _i,  i=1, 2$ are the two regions partitioned by the curve $C$ and they change as the curve evolves. Since we are able to construct regions $\Omega_i$ from an open contour $C$, it is easy to calculate $u_i$ in (\ref{eq:statisticsCal}). After $u_i$ are estimated, we fix $u_i$ and use the gradient descent method to minimise the functional (\ref{eq:ourModel}) with respect to the open contour $C$. This results in the following contour evolution equation
\begin{equation} \label{eq:grdientDecofOurModel}
\frac{{\partial C\left( {s,t} \right)}}{{\partial t}}  = - \frac{{\partial E(C)}}{{\partial C}} = \alpha {C_{ss}}\left( {s,t} \right) - \beta {C_{ssss}}\left( {s,t} \right) - Q\left( {s,t} \right) \boldsymbol{n}\left( {s,t} \right).
\end{equation}
where $Q(s,t) = \left( {f - {u_1}} \right)^2 - \left( {f - {u_2}} \right)^2$, and $\boldsymbol{n}(s,t)=(\boldsymbol{n}_x(s,t),\boldsymbol{n}_y(s,t))$ denotes the outer normal vector of the contour $C$. The first two terms on the right-hand side of (\ref{eq:grdientDecofOurModel}) minimise the contour length and thereby enforce an equidistant spacing between the control points. The third term maximises the homogeneity in the adjoining regions in the narrowband, which is measured by the energy density (\ref{eq:statisticsCal}). This forces $C$ move towards a retinal layer boundary in the OCT image.

We now need to discretise (\ref{eq:grdientDecofOurModel}) for numerical implementation. Specifically, for each control point of the contour $C$, we have the following two semi-implicit iterative schemes
\begin{equation} \label{eq:grdientDecXofOurModel}
\frac{{x_i^{k + 1} - x_i^k}}{{\Delta t}} = \alpha {\partial _{xx}}x_i^{k + 1} - \beta {\partial _{xxxx}}x_i^{k + 1} - Q(x_i^k,y_i^k){\boldsymbol{n}_x}(x_i^k,y_i^k),
\end{equation}
\begin{equation} \label{eq:grdientDecYofOurModel}
\frac{{y_i^{k + 1} - y_i^k}}{{\Delta t}} = \alpha {\partial _{yy}}y_i^{k + 1} - \beta {\partial _{yyyy}}y_i^{k + 1} - Q(x_i^k,y_i^k){\boldsymbol{n}_y}(x_i^k,y_i^k),
\end{equation}
where $\Delta t$ is an artificial time parameter. The superscript $k$ denotes the $k$th iteration, and the subscript $i$ the $i$th control point. As compared to explicit iterative methods, semi-implicit relaxation methods allow the use of a relative larger time step ${\Delta t}$, speeding up the convergence rate of contour evolution. We apply the finite difference method with Neumann boundary condition {{\cite{Noye1990Accurate}}} to discretise the 2nd- and 4th-order derivatives $\partial _{xx}x_i$, $\partial _{yy}y_i$, $\partial _{xxxx}x_i$ and $\partial _{yyyy}y_i$.

To iterate (\ref{eq:grdientDecXofOurModel}) and (\ref{eq:grdientDecYofOurModel}), the open contour normals $({\boldsymbol{n}_x}, {\boldsymbol{n}_y})$ are calculated in Fig~\ref{fig:contourNormals}. Apart from the two endpoints, for each control point on the contour we first compute its tangent using its neighbours via the centre finite difference scheme. For the start and end points, their tangents are respectively computed with the forward and backward finite differences. The contour normals thereby can be easily derived from the computed tangents because they are perpendicular to each other.

Note that given $M$ control points we have to calculate a set of 2$M$ equations as in (\ref{eq:grdientDecXofOurModel}) and (\ref{eq:grdientDecYofOurModel}).
Since the 2nd- and 4th-order derivatives are linear differential operators, we can rewrite (\ref{eq:grdientDecXofOurModel}) and (\ref{eq:grdientDecYofOurModel}) to its matrix form as
\begin{equation}\label{eq:X}
{\vect{x}^{k + 1}} = {\left( {I - \Delta t\textbf{A}} \right)^{ - 1}}\left( {{\vect{x}^k} - \Delta tQ\left( {{\vect{x}^k},{\vect{y}^k}} \right) \cdot {\boldsymbol{n}_x}\left( {{\vect{x}^k},{\vect{y}^k}} \right)} \right),
\end{equation}
\begin{equation}\label{eq:Y}
{\vect{y}^{k + 1}} = {\left( {I - \Delta t\textbf{A}} \right)^{ - 1}}\left( {{\vect{y}^k} - \Delta tQ\left( {{\vect{x}^k},{\vect{y}^k}} \right) \cdot {\boldsymbol{n}_y}\left( {{\vect{x}^k},{\vect{y}^k}} \right)} \right),
\end{equation}
where $I$ is the identity matrix, $\cdot$ denotes pointwise multiplication and $\textbf{A}$ is an $M \times M$ matrix of the form

\begin{equation*} \nonumber
\resizebox{0.8\textwidth}{!}{$
\left( {\begin{array}{*{20}{c}}
{ - \alpha  - 2\beta }&{\alpha  + 3\beta }&{-\beta }&0&0&0& \cdots &0&0\\
{\alpha  + 3\beta }&{ - 2\alpha  - 6\beta }&{\alpha  + 4\beta }&{-\beta }&0&0& \cdots &0&0\\
{-\beta }&{\alpha  + 4\beta }&{ - 2\alpha  - 6\beta }&{\alpha  + 4\beta }&{-\beta }&0& \cdots &0&0\\
0&{-\beta }&{\alpha  + 4\beta }&{ - 2\alpha  - 6\beta }&{\alpha  + 4\beta }&{-\beta }& \cdots &0&0\\
0&0&{-\beta }&{\alpha  + 4\beta }&{ - 2\alpha  - 6\beta }&{\alpha  + 4\beta }& \cdots &0&0\\
 \vdots & \vdots & \vdots & \vdots & \vdots & \vdots & \ddots & \vdots & \vdots \\
0&0&0&0&0&0& \cdots &{ - 2\alpha  - 6\beta }&{\alpha  + 3\beta }\\
0&0&0&0&0&0& \cdots &{\alpha  + 3\beta }&{ - \alpha  - 2\beta }
\end{array}} \right)$}.
\end{equation*}
Matrix $\textbf{A}$ satisfies the condition that $C(0) \ne C(1)$. It is a sparse matrix and only 5 diagonals have non-zero values so can be inverted efficiently. The equations (\ref{eq:X}) and (\ref{eq:Y}) are now discretised with a set of control points. The solutions gives the coordinates of all the control points $(\vect{x},\vect{y})$ at $(k+1)$th iteration. \\
\begin{figure}
\vspace{-20pt}
\centering
{\includegraphics[width=0.7\textwidth]{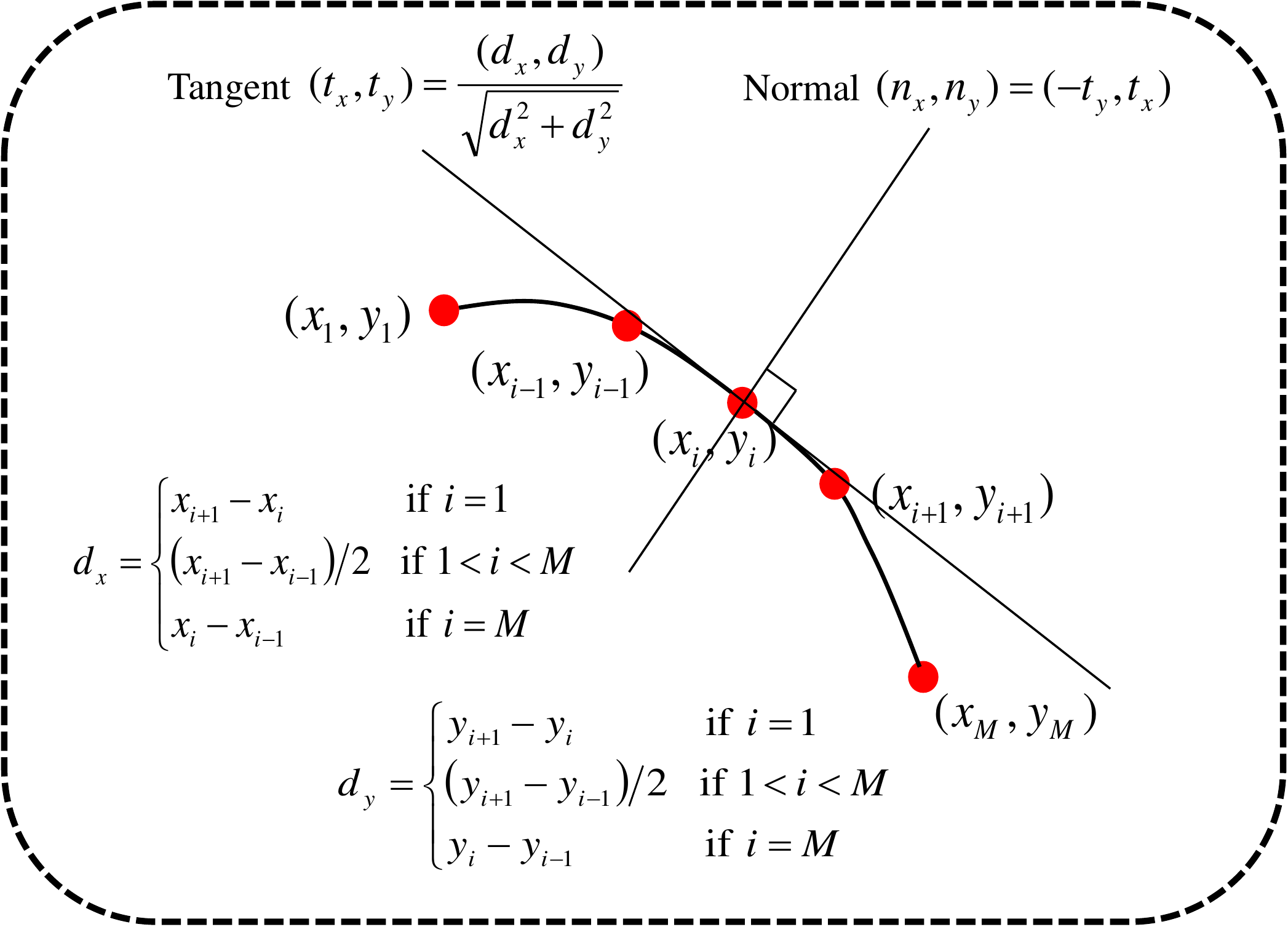}}
\caption{Computing normals of a discrete point on an open parametric curve.}
\label{fig:contourNormals}
\vspace{-20pt}
\end{figure}

\subsection{Integrating shape constraint to intensity-based MS}
The contour evolution driven by (\ref{eq:X}) and (\ref{eq:Y}) can become irregular as the MS functional uses only local pixel intensities. To further improve the MS functional we propose to impose some constraint for the functional to control contour evolution. With some manually segmented retinal layer boundaries as training shapes we can derive the statistical distribution of the retinal layer boundaries as a shape constraint on the MS functional. The main idea of this is to project the irregular shapes into a latent space spanned {by a few eigenvectors of training shapes with largest eigenvalues}. In this way, the resulting irregular shapes are restricted to a manifold of familiar shapes and thereby can be corrected. Note that other dimensionality reduction methods can be also used to derive shape constraint. In this work we found PCA performs reasonably well. Next we detail the use of shape constraint to (\ref{eq:ourModel}) using PCA. Note that an OCT shape model represents the locations of 9 boundaries.

If one OCT shape model consists of $N$ control points, it can be modelled as a $2N$-dimensional vector $\vect{z} = {\left( {{x_1}, \cdots, {x_N},{y_1},\cdots,{y_N}} \right)^T}$, where $N=9M$ ($M$ appears in (\ref{eq:ctrpitvector})). Assuming that we have $L$ training shapes manually annotated from $L$ OCT images. We first align all shapes at the centre of coordinate origin by applying {{the Procrustes} transformations. The mean OCT shape is then computed by $\tilde {\vect{z}} = \tfrac{1}{L}\sum\nolimits_{j = 1}^L {{\vect{z}_j}}$. For each shape $\vect{z}_j$ in the training set, its deviation from the mean shape $\tilde {\vect{z}}$ is $\vect{dz}_j = \vect{z}_j - \tilde {\vect{z}}$. Then the {{$2N \times 2N$} covariance matrix $\textbf{Cov}$ can be calculated by $\textbf{Cov} = \frac{1}{L-1}\sum\nolimits_{j = 1}^L {\vect{dz}_j} \vect{dz}_j^T$.

We calculate the eigenvectors $\vect{v}_k$ and corresponding eigenvalues $\lambda_k$ of $\textbf{Cov}$ (sorted so that $\lambda_k \le \lambda_{k+1}$). Let $\textbf{P} = \left( {{\vect{p}_1}, \ldots ,\vect{p}_m} \right)$ be the matrix of the first $m$ eigenvectors of $\vect{v}_k$. Then we can approximate an OCT shape in the training set by
\begin{equation} \label{eq:PCA}
\vect{z} {{\approx}} \tilde{\vect{z}} + \textbf{P}\vect{b},
\end{equation}
where $\vect{b} = \left( {{b_1}.b_2, \ldots ,b_m} \right)$ defines the parameters {{for $m$} different deformation patterns. Since $\textbf{P}$ is an orthogonal matrix, $\vect{b}$ is given by
\begin{equation} \label{eq:inverseB}
\vect{b} = \textbf{P}^T(\vect{z} - \tilde{\vect{z}}).
\end{equation}
By varying the parameters $b_i$, we can generate new examples of the shape. We can also limit each $b_i$ to constrain the deformation patterns of the shape. Typical limits \cite{cootes2000introduction} are
\begin{equation} \label{eq:constrainB}
- 3\sqrt {{\lambda _i}}  \le {b_i} \le 3\sqrt {{\lambda_i}},
\end{equation}
where $i =  1,\ldots ,m$.

The use of shape constraint to (\ref{eq:ourModel}) for segmenting 9 retinal layer boundaries is summarised as follow: $\textbf{P}$, $\tilde {\vect{z}}$ and $\lambda_i$ are first computed from a set of training OCT shapes. After the original OCT shape model (locations of 9 boundaries) is updated with (\ref{eq:X}) and (\ref{eq:Y}) (each boundary in the shape model is updated separately), it is {{transformed to approximate the mean shape $\tilde {\vect{z}}$ with the Procrustes transformation}, which forms ${\vect{z}}$. This is followed by updating $\vect{b}$ using (\ref{eq:inverseB}) and then projecting it using (\ref{eq:constrainB}){{, i.e. the coefficients $b$ obtained with (\ref{eq:inverseB}) are constrained with equation \eqref{eq:constrainB}.}} Next, we correct the original OCT shape using (\ref{eq:PCA}) with the projected $\vect{b}$ and warping the corrected version back to the original location. In this way, the original irregular OCT shape is pulled back to a regular one used for the next iteration of (\ref{eq:X}) and (\ref{eq:Y}). The whole process is repeated until convergence. The overall numerical optimisation of the proposed method is fast as it only computes a few hundred control points.


\section{Experimental results}
\label{sec:experiments}

\textbf{Data}: 9 retinal layers boundaries were segmented for each image from in vivo OCT B-scans. 30 Spectralis SDOCT (ENVISU C class 2300, Bioptigen, axial resolution = 3.3$\mu m$, scan depth = 3.4$mm$, 32,000 A-scans per second) B-scans from 15 healthy adults were used for the work. The B-scan was imaged from the left and right eye of the healthy adults using a spectral domain OCT device with a chin rest to stabilise the head. The B-scan located at the foveal centre was identified from the lowest point in the foveal pit where the cone outer segments were elongated (indicating cone specialisation). The ground truth boundaries were manually generated by an experienced ophthalmologist. The dataset was randomly split into 20 training and 10 (5 are corrupted with high-level speckle noise using Matlab $imnoise$ function with 0.8 variance) validation datasets. For image pre-processing, all images were cropped to extract only region of interest and were flattened using ground truth labels before training. The 9 retinal layer boundaries which can be segmented by the proposed method are shown in Fig~\ref{fig:OCTBoundary}.
\begin{figure}[h!] 
\vspace{-20pt}
\centering  
{\includegraphics[height=0.23\textwidth]{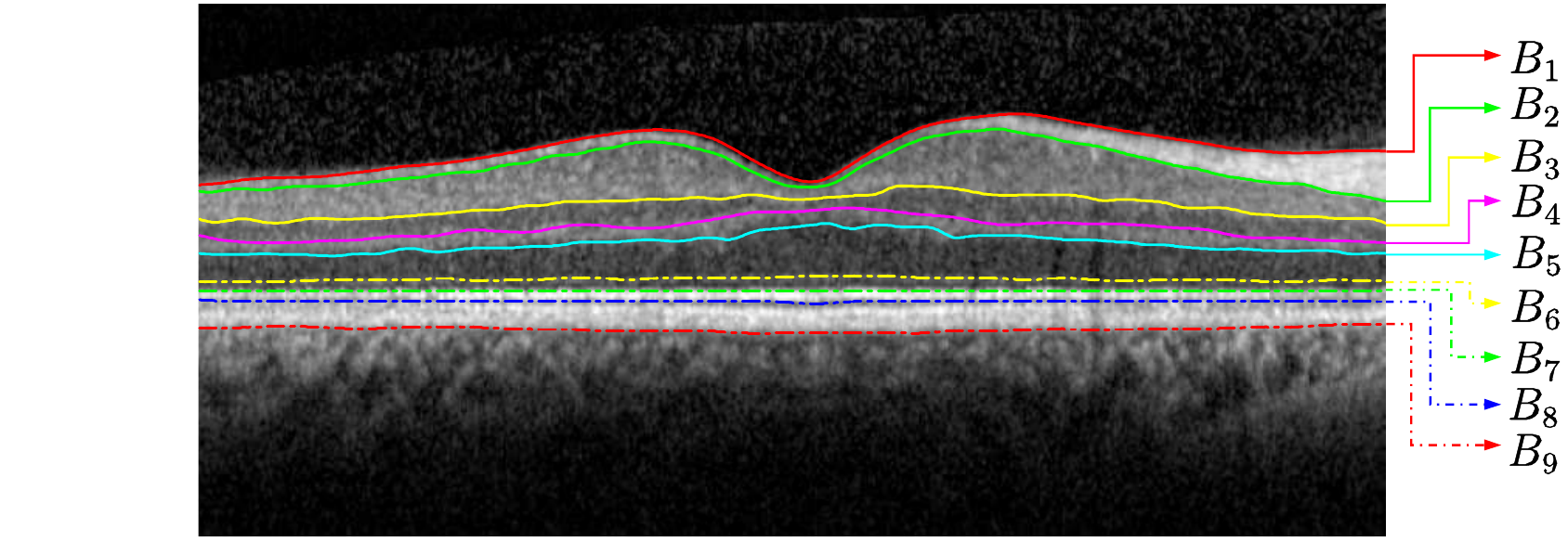}}
\vspace{-5pt}
\caption{An example B-Scan OCT image centred at the macula, showing 9 target intra-retinal layer boundaries. The names of these boundaries labelled as notations $B_1$,$B_2$...$B_9$ are summarised in Table~\ref{tb:OCTBoundary}.}
\label{fig:OCTBoundary}
\vspace{-40pt}
\end{figure}

\begin{table}[h!]
\centering  
\caption{Notations for nine retinal boundaries, their corresponding names and abbreviations}
\resizebox{\columnwidth}{!}{
\begin{tabular}{lcc}
\toprule
Notation  & Name of retinal boundary/surface & Abbreviation\\
\midrule
$B_1$&  internal limiting membrane & ILM \\
$B_2$&  outer boundary of the retinal nerve fibre layer & RNFL$_o$\\
$B_3$&  inner plexiform layer-inner nuclear layer & IPL-INL  \\
$B_4$&  inner nuclear layer-outer plexiform layer & INL-OPL \\
$B_5$&  outer plexiform layer-outer nuclear layer & OPL-ONL \\
$B_6$& 	outer nuclear layer-inner segments of photoreceptors & ONL-IS\\
$B_7$&  inner segments of photoreceptors-outer segments of photoreceptors &IS-OS \\
$B_8$&  outer segments of of photoreceptors-retinal pigment epithelium & OS-RPE\\
$B_9$&  retinal pigment epithelium-choroid & RPE-CH\\
\bottomrule
\end{tabular}}
\label{tb:OCTBoundary}
\vspace{-15pt}
\end{table} 

\noindent \textbf{Parameters}: These parameters in our method are the first and second order regularisation parameters $\alpha$ and $\beta$, the artificial time step $\Delta t$, the narrowband radius $|s|$, the number of eigenvectors/eigenvalues $t$, the number of control points for 9 boundaries, and the iteration number (1000).

These parameters are selected as follows: 1) Large $\alpha$ and $\beta$ leads to increasingly shorter and finally vanishing segmentation boundaries. Small values may cause control points intersect with each other, thus leading to numerical instabilities. In this work, we fixed $\alpha=\beta=0.5$; 2) $\Delta t$ is bounded by the CFL stability condition. Numerical stabilities can be attained by using $\Delta t = 0.01$; 3) $|s|$ was selected according to the initial OCT boundaries. If initialisation is close to the true retinal layer boundaries, $|s|$ is small ($10$ {{pixels} for Fig~\ref{fig:initial} 1st row). Otherwise, a large $|s|$ should be used ($50$ {{pixels} for Fig~\ref{fig:initial} last row). Note that we set $|s|$ the same value for each of 9 OCT boundaries; 4) $t$ was confirmed by choosing the first $t$ largest eigenvalues such that $\sum\nolimits_{i = 1}^t {{\lambda _i}}  \ge 0.98{V_T}$, where ${V_T}$ is the total variance of all the eigenvalues. We used the number of training samples as the value of $t$. Eigenvectors corresponding to small eigenvalues do not contribute much to shape variation; 5) 360 control points in total are used for a whole OCT shape (each retinal layer boundary thus has 40). Overall, we only adjusted $|s|$ for different initialisations (see Fig~\ref{fig:initial}).

\begin{figure}[h!]
\centering
\vspace{-10pt}
{\includegraphics[width=0.89\textwidth]{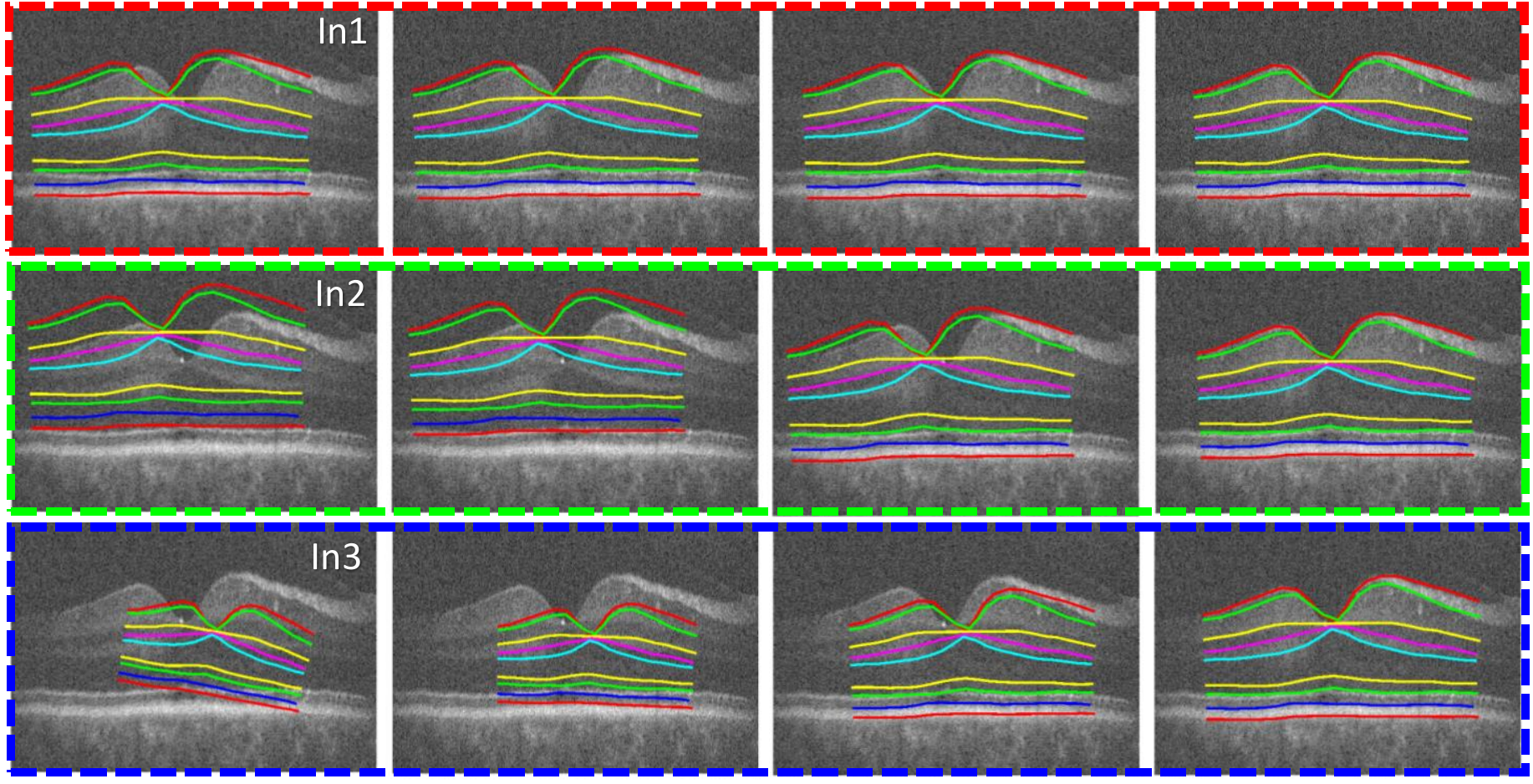}}\\
\vspace{-5pt}
\caption{Experiments of contour evolution with different initialisation conditions for the proposed method. 1st column: three initialisations (i.e. In1, In2 and In3); 2nd-3rd columns: intermediate contour evolution; 4th column: final segmentation results; 5th column: notation abbreviations for 9 retinal layer boundaries (refer to Fig~\ref{fig:OCTBoundary} for their full name). The experiments show that by incorporating region-based information our method has much larger convergence range so it is robust against different initialisations.}
\label{fig:initial}
\vspace{-15pt}
\end{figure}

\noindent \textbf{Comparsion}: The performance of the proposed segmentation method was evaluated by computing the Hausdorff distance (HD) metric between the automated and ground truth segmentations for different retinal layer boundaries. We compared our method with the geodesic distance method proposed in \cite{duan2017automated}. In Fig~\ref{fig:visual}, visual comparison suggests that the proposed method provides significant improvements over intensity-based geodesic distance in OCT segmentation, especially when OCT images are of low contrast and contain high-level speckle noise. In Table~\ref{tb:numb}, we report the HD metric of each boundary as well as the total 9 boundaries over the validation dataset and show that our method outperforms the geodesic distance in terms of the HD metric for all the retinal layer boundaries. The improvements are more evident at IML, RNFL$_\sigma$, IPL-INL, INL-OPL and OPL-ONL boundary locations. Note that in order to compare our results with ground truth labels we fitted a spline curve to each segmentation contour such that the resulting contours run across the entire width of the OCT image.
\begin{figure}[h!]
\vspace{-10pt}
\centering
{\includegraphics[width=0.89\textwidth]{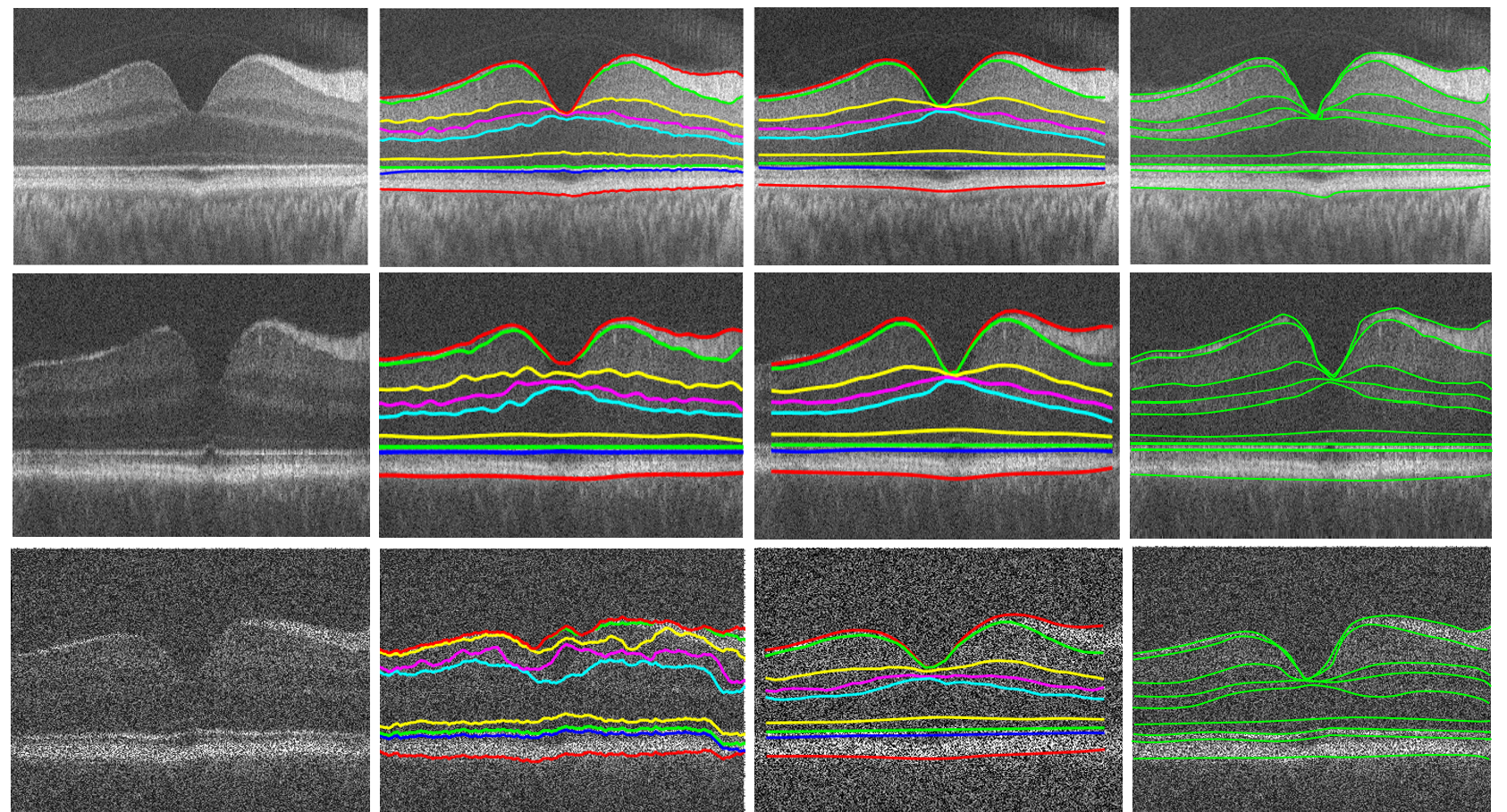}}\\
\vspace{-5pt}
\caption{Visual comparison of segmentation of 9 retinal layer boundaries using the geodesic distance \cite{duan2017automated} and proposed shape-based MS method. 1st column: input B-scans; 2nd column: geodesic distance results; 3rd column: our results; 4th column: ground truth.}
\label{fig:visual}
\vspace{-20pt}
\end{figure}

\begin{table}[h!]
\vspace{-20pt}
\centering
\caption{Quantitative comparison of segmentation results from the geodesic distance method (GDM) and proposed method for different retinal layer boundary, in terms of Hausdorff distance metric (mean$\pm$standard deviation).}
\begin{tabular}{|c|c|c|c|c|c|} \hline
{}            & IML & NFL$_\sigma$ & IPL-INL & INL-OPL & OPL-ONL     \\ \hline
GDM \cite{duan2017automated}           &9.80$\pm$2.51  &10.2$\pm$3.32 &25.26$\pm$10.5 &22.55$\pm$8.96 &20.46$\pm$7.24     \\\hline
Proposed          &1.72$\pm$0.52  &1.68$\pm$0.66 &1.058$\pm$0.12 &0.925$\pm$0.10 &0.863$\pm$0.23     \\ \hline
{}            & ONL-IS & IS-OS & OS-RPE & RPE-CH & Overall    \\ \hline
GDM \cite{duan2017automated}          &2.53$\pm$1.05 &1.91$\pm$1.34 &1.21$\pm$1.05 &1.031$\pm$0.98 &10.55$\pm$4.10    \\\hline
Proposed          &0.61$\pm$0.05 &0.84$\pm$0.11 &1.61$\pm$0.96 &0.952$\pm$0.78 &1.140$\pm$0.39    \\ \hline
\end{tabular}
\label{tb:numb}
\vspace{-15pt}
\end{table}
\section{Conclusion}
\label{sec:conclusion}
We presented a new segmentation method for optical coherence tomography (OCT) images, which allows the integration of statistical shape models learned from a small OCT dataset. To this end, we developed the Mumford-Shah functional in a way which facilitates a parametric representation of open contours. We then constructed narrowbands around the open contours such that regional information can be derived to assist segmentation. We have shown that integrating such information allows the proposed method to have a large convergence range and thus robust against different initialisations. We have also validated that the proposed method is very accurate even OCT images are of low contrast and contain high-level speckle noise, and that the method outperforms the state-of-the-art geodesic distance segmentation method. The proposed method can be readily extended to other segmentation problems involving open contours.

\bibliographystyle{splncs}
\bibliography{cites}

\end{document}